\documentclass[lettersize,journal]{IEEEtran}
\usepackage{amsmath,amsfonts}
\usepackage{algorithmic}
\usepackage{algorithm}
\usepackage{array}
\usepackage[caption=false,font=normalsize,labelfont=sf,textfont=sf]{subfig}
\usepackage{textcomp}
\usepackage{stfloats}
\usepackage{url}
\usepackage{verbatim}
\usepackage{graphicx}
\usepackage{cite}
\usepackage{amsmath}
\usepackage{amsfonts}
\usepackage{bm}
\usepackage{graphicx}
\usepackage{multirow}
\usepackage{booktabs}
\usepackage{color}
\begin{document}

\title{Multiscale Latent-Guided Entropy Model for LiDAR Point Cloud Compression}

\author{Tingyu Fan, Linyao Gao, Yilin Xu, Dong Wang,~\IEEEmembership{Member,~IEEE,} Zhu Li,~\IEEEmembership{Senior Member,~IEEE}

}

\markboth{IEEE TRANSACTIONS ON CIRCUITS AND SYSTEMS FOR VIDEO TECHNOLOGY,~Vol.~14, No.~8, Jan~2023}%
{Shell \MakeLowercase{\textit{et al.}}: A Sample Article Using IEEEtran.cls for IEEE Journals}


\maketitle

\begin{abstract}
The non-uniform distribution and extremely sparse nature of the LiDAR point cloud (LPC) bring significant challenges to its high-efficient compression. This paper proposes a novel end-to-end, fully-factorized deep framework that represents the original LiDAR point cloud into an octree structure and hierarchically constructs the octree entropy model in layers. The proposed framework utilizes a hierarchical latent variable as side information to encapsulate the sibling and ancestor dependence, which provides sufficient context information for the modeling of point cloud distribution while enabling the parallel encoding and decoding of octree nodes in the same layer. Besides, we propose a residual coding framework for the compression of the latent variable, which explores the spatial correlation of each layer by progressive downsampling, and model the corresponding residual with a fully-factorized entropy model. Furthermore, we propose soft addition and subtraction for residual coding to improve network flexibility. The comprehensive experiment results on the LiDAR benchmark SemanticKITTI and MPEG-specified dataset Ford demonstrate that our proposed framework achieves state-of-the-art performance among all the previous LPC frameworks. Besides, our end-to-end, fully-factorized framework is proved by experiment to be high-parallelized and time-efficient, which saves more than 99.8\% of decoding time compared to previous state-of-the-art methods on LPC compression.
\end{abstract}

\begin{IEEEkeywords}
point cloud compression, end-to-end learning, octree, deep entropy model, sparse convolution.
\end{IEEEkeywords}

%
\IEEEpeerreviewmaketitle

\section{Introduction} \label{intro}

\IEEEPARstart{D}{ue} to the rapid development of 3D sensors, point clouds (PC) \cite{xu2018introduction, zhu2020view, zhu2021lossy} have become a promising data structure with broad applications in autonomous driving \cite{akhtar2019low}, robotic sensing \cite{zhang2021attan}, and AR/VR. Among various use cases, a type of point cloud captured by the Light Detection And Ranging (LiDAR) sensor, called the LiDAR point cloud (LPC) \cite{9906120}, has recently received much attention. An LPC represents the scene information centered on a driving vehicle and is usually collected in real-time, with over 100,000 points per frame \cite{li2022frame}. However, the non-uniform distribution of the LiDAR point cloud and the noise brought by real-time acquisition make the exploration of spatial correlation extremely difficult, bringing significant challenges to its compression. Therefore, this paper focuses on the geometry of LiDAR point clouds, aiming to design an efficient end-to-end deep framework for the compression of LiDAR point clouds.

Prior works on LiDAR point cloud compression use multiple data structures to regularize the non-uniform point cloud sequence, e.g., range images \cite{houshiar20153d}, KD-trees \cite{bentley1975multidimensional,devillers2000geometric} and Octrees \cite{schnabel2006octree, meagher1982geometric}. Among them, the Moving Picture Expert Group (MPEG) develops a Geometry-based Point Cloud Compression (G-PCC) standard \cite{schwarz2018emerging}. G-PCC is implemented based on the octree structure that recursively divides the current 3D cube into eight sub-cubes. G-PCC reports state-of-the-art performance among all rule-based methods. However, the above methods are highly dependent on hand-crafted context selection and entropy model without the optimization by traversing the whole dataset, leading to unsatisfactory coding efficiency.

Thanks to the application of deep learning techniques in image/video compression \cite{balle2017end,balle2018variational,lu2019dvc,hu2021fvc,hu2022coarse,li2021deep}, learning-based point cloud compression methods \cite{huang2020octsqueeze,biswas2020muscle,que2021voxelcontext,fu2022octattention} have emerged. For LPC, most of the methods are based on the octree entropy model. Given the octree sequence ${\bf x}=\{x_1, x_2,\cdots,x_N\}$, the lower bound of bit rate can be formulated using the information entropy of the data distribution $p({\bf x})$ \cite{shannon2001mathematical}, which is assumed intractable due to its high dimensionality. Therefore, the goal is to approximate $p({\bf x})$ with an estimated distribution $q({\bf x})$ computed by a deep neural network. Given an estimation $q({\bf x})$, the lower bound of bit rate is the cross-entropy between $q({\bf x})$ and $p({\bf x})$, i.e.,
\begin{equation}
H(p,q)=\mathbb{E}_{{\bf x}\sim p}\left[-\log_2 q({\bf x})\right],
\end{equation}
which is a tight lower bound achievable by arithmetic coding algorithms. Due to the high-dimensionality of $\bf x$, the probability model $q({\bf x})$ is often constructed in an auto-regressive \cite{van2016pixel, salimans2017pixelcnn++} fashion:
\begin{equation}
    q({\bf x})=\prod_i^N q(x_i|{\rm context}(x_i)),
\end{equation}
where ${\rm context}(x_i)$ is a subset of all decoded nodes that are considered to have dependence with $x_i$. The strategy to determine the context set varies among different methods. OctSqueeze \cite{huang2020octsqueeze} assumes the conditional independence of sibling nodes given their ancestors and applies a hierarchical coding structure that selects ancestry nodes as context, bringing low computational complexity at the cost of violating the original distribution. To mitigate the huge information loss brought by the sibling independence assumption, VoxelContext-Net \cite{que2021voxelcontext} utilizes a 3D convolutional neural network (CNN) to capture the dependencies between nodes in the bottom layers of the octree. OctAttention \cite{fu2022octattention} further encodes and decodes the octree in a breadth-first order and maintains a context window $\{x_{i-N+1},\cdots,x_{i-1}\}$ for attention-based context learning, which reports state-of-the-art performance among all LPC compression methods. However, selecting the decoded sibling nodes causes the context to vary during the decoding process. Thus the model needs to be constantly re-run to update the semantic information of the current context, leading to excessive decoding complexity and non-parallelism. Although the author proposes mask operation for complexity reduction, the model still requires even more than 10 minutes to decode one frame of point cloud \cite{fu2022octattention}.

\begin{figure*}[htbp]
\centering  
\includegraphics[width=0.99\textwidth]{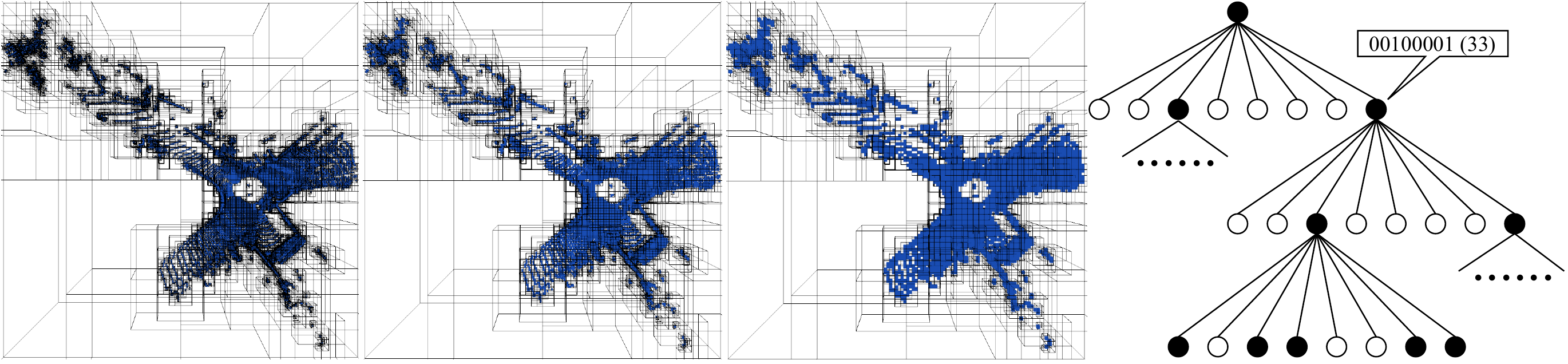}
\caption{The demonstration of octrees. The left three graph shows the hierarchical construction process of the octree sequence with depths 9, 8, and 7. The rightmost graph shows a part of an octree sequence, where black indicates occupied, and white indicates unoccupied. The occupied nodes need to be subdivided into smaller cubes until reaching the specified depth. }
\label{pic_octree}
\end{figure*}

To summarize, two major challenges must be compatibly addressed to design an efficient LPC compression network: 1) The context selection algorithm needs to thoroughly explore the dependence of the whole octree to avoid suboptimality. 2) The decoding process of nodes should be independent enough, such that the model runs only {\bf once} during decoding for the sake of complexity and parallelism.

To this end, we propose a new end-to-end, fully-factorized LPC entropy model. Inspired by Ball\'e et al. \cite{balle2018variational}, we address the above two issues by transmitting {\it side information}, which encodes additional information to reduce the mismatch of the entropy model. The proposed {\it side information} is a latent variable encapsulating the dependency information of the octree, through which the octree nodes are assumed to be independent layer-wisely. To fully explore the spatial correlation, we design the latent variable in a hierarchical form, progressively increasing the perceptive field to capture deeper correlation information. The model we propose encapsulates the sibling correlation through the latent variable to achieve optimal coding efficiency. Meanwhile, compared with Fu et al., the independence of sibling nodes conditioned on the latent variable significantly decreases the computational complexity and brings decoding parallelism among nodes in the same layer, such that the model runs only {\bf once} during decoding. To summarize, our contributions are highlighted as follows:

\begin{itemize}
    \item We propose an end-to-end, fully-factorized LPC compression framework that encapsulates the spatial correlation of point clouds by introducing a hierarchical latent variable. The proposed framework explores the dependence of sibling nodes to avoid suboptimality while maintaining low complexity and a high degree of parallelism. The model runs only {\bf once} during decoding.

    \item We propose to replace the ordinary addition/subtraction operation with soft addition/subtraction to improve the model's flexibility.
    
    \item The experiment result shows that our proposed network achieves state-of-the-art performance on LPC datasets SemanticKitti and Ford. Furthermore, due to the enormous complexity reduction, our model saves 99.8\% of runtime compared with the previous state-of-the-art method.
\end{itemize}

\section{RELATED WORK}
\subsection{Sparse Convolution}
Compared with well-structured 2D images/videos, point clouds are of unordered and non-uniform characteristics. A straightforward way to process 3D point clouds is by voxelization and 3D convolutional neural networks (CNN), which runs counter to the sparsity nature of point clouds, resulting in huge time and memory consumption. Therefore, Choy et al. propose the generalized sparse convolution \cite{choy20194d}, with the original point cloud $\mathcal{P}=\{(x_i,y_i,z_i,{\bf f_i})\}_i$ expressed as the coordinate matrix $C$ and the associated feature matrix $F$:
\begin{equation}
C=\begin{bmatrix}
 b_1 & x_1 & y_1 & z_1\\ 
 b_2 & x_2 & y_2 & z_2\\
     &     & \vdots & \\
 b_N & x_N & y_N & z_N
 \end{bmatrix},
 F=\begin{bmatrix}
 {\bf f_1}\\ 
 {\bf f_2}\\
  \vdots \\
  {\bf f_N}
 \end{bmatrix}
\end{equation}
where $b_i$ is the batch index and $\bf f_i$ is the associated feature. The generalized sparse convolution is defined as:
\begin{equation}
x_u^{out}=\sum_{i\in \mathcal{N}^3( u,\mathcal{C}^{in})}W_ix_{u+i}^{in} \ {\rm for}\ u\in \mathcal{C}^{out}
\end{equation}
where $x_{u}^{in}$ is the input feature-vector defined at $u$, $\mathcal{C}^{in}$ and $\mathcal{C}^{out}$ are predefined input and output coordinates and $\mathcal{N}^3(u,\mathcal{C}^{in})=\{i|u+i\in \mathcal{C}^{in},i\in \mathcal{N}^3\}$ is the set of offsets from $\bf u$ that exist in $\mathcal{C}^{in}$. The sparse CNN utilizes the same convolutional structure as the dense CNN for each effective point while significantly reducing the time and memory consumption.
\subsection{Rule-based LiDAR Point Cloud Compression}

Tree structures are widely applied in LPC compression due to its ability to regularize sparse geometry data, e.g., KD-tree \cite{devillers2000geometric}, quadtree \cite{kathariya2018scalable} and octree \cite{schnabel2006octree,schwarz2018emerging}. Among them, the octree-based standard G-PCC \cite{schwarz2018emerging}  developed by the MPEG group achieves state-of-the-art performance. G-PCC relies on the decomposition of octree geometry. Specifically, a LiDAR point cloud $\bf P$ is first quantized into a $2^L\times2^L\times2^L$ cube:
\begin{equation}
\begin{split}
    &\hat{\bf P}={\rm round}\left(\frac{{\bf P}-{\rm bias}}{\rm qs}\right),
    \\
    &{\rm qs}=\frac{\rm bounding}{2^L-1},
\end{split}
\end{equation}
where ${\rm bias}=[{\rm min}({\bf P}_x), {\rm min}({\bf P}_y), {\rm min}({\bf P}_z)]$, $L$ is the target depth of the octree, and ${\rm bounding}$ is the size of the bounding box of $\bf P$. Correspondingly, the inverse quantization is defined as follows:
\begin{equation}
    \bar{\bf P}=\hat{\bf P}*{\rm qs}+{\rm bias},
\end{equation}
where $\bar{\bf P}$ is the reconstructed point cloud. As Figure \ref{pic_octree} shows, the octree is constructed by recursively dividing the current cube of $\hat{\bf P}$ into 8 subcubes until the current cube is empty or the recursion reaches the target depth $L$. An 8-bit occupancy code is assigned for each cube to indicate the occupancy situation (0 for unoccupied and 1 for occupied). However, the above methods (including G-PCC) rely on the hand-crafted context model that cannot be optimized by traversing large-scale data, leading to inferior performance compared with deep-learning based methods.

\subsection{Learning-based Point Cloud Compression}
Due to the applications of deep learning techniques in image/video compression, learning-based point cloud compression methods have recently emerged. In general, these learning-based methods can be summarized as point-based, voxel-based and octree-based.
\subsubsection{Point-based methods}
The point-based methods directly process and compress the original point cloud, e.g., Huang et al. \cite{huang20193d} utilizes Pointnet++ \cite{qi2017pointnet++} structure for feature extraction and point cloud reconstruction. Gao et al. \cite{gao2021point} further propose graph-based feature extraction and attentive sampling to improve compression efficiency. However, due to the complexity factor, the point-based methods are limited to small-scale datasets like ShapeNetCorev2 \cite{chang2015shapenet}.

\subsubsection{Voxel-based methods}
The voxel-based methods process the point clouds by voxelization. Among them, Wang et al. \cite{wang2021lossy} utilize a 3D CNN-based network for voxelized point cloud compression. The experiment result shows remarkable coding gain; however, the voxelization process runs counter to the sparsity nature of point clouds, leading to excessive time and memory consumption. Therefore, Wang et al. further utilize the generalized sparse convolution by Choy et al. \cite{choy20194d} and propose an end-to-end multi-scale point cloud compression framework based on the sparse CNN \cite{wang2021multiscale}. The experiment result shows state-of-the-art performance on dense point cloud datasets 8i Voxelized Full Bodies \cite{d20178i} (8iVFB) and Microsoft Voxelized Upper Bodies (MVUB) \cite{d2016mvub}. However, the above voxel-based methods are limited to dense point clouds, which are not yet implementable on LPC \cite{wang2021lossy,wang2021multiscale} due to the extreme sparsity and noise brought by real-time acquisition.

\subsubsection{Octree-based methods}
As mentioned in Section \ref{intro}, The octree-based methods utilize the octree structure to regularize the point cloud structure, which is widely used in LiDAR point cloud compression due to the ability to express sparse point clouds. The auto-regressive probability model is widely used to infer the high-dimensional octree sequence. However, due to the inefficiency of the auto-regressive inference, existing octree-based methods \cite{huang2020octsqueeze,biswas2020muscle,que2021voxelcontext,fu2022octattention} fail to balance the performance and complexity (more details in Section \ref{intro}).

\begin{figure}[t]
\centering  
\includegraphics[width=0.495\textwidth]{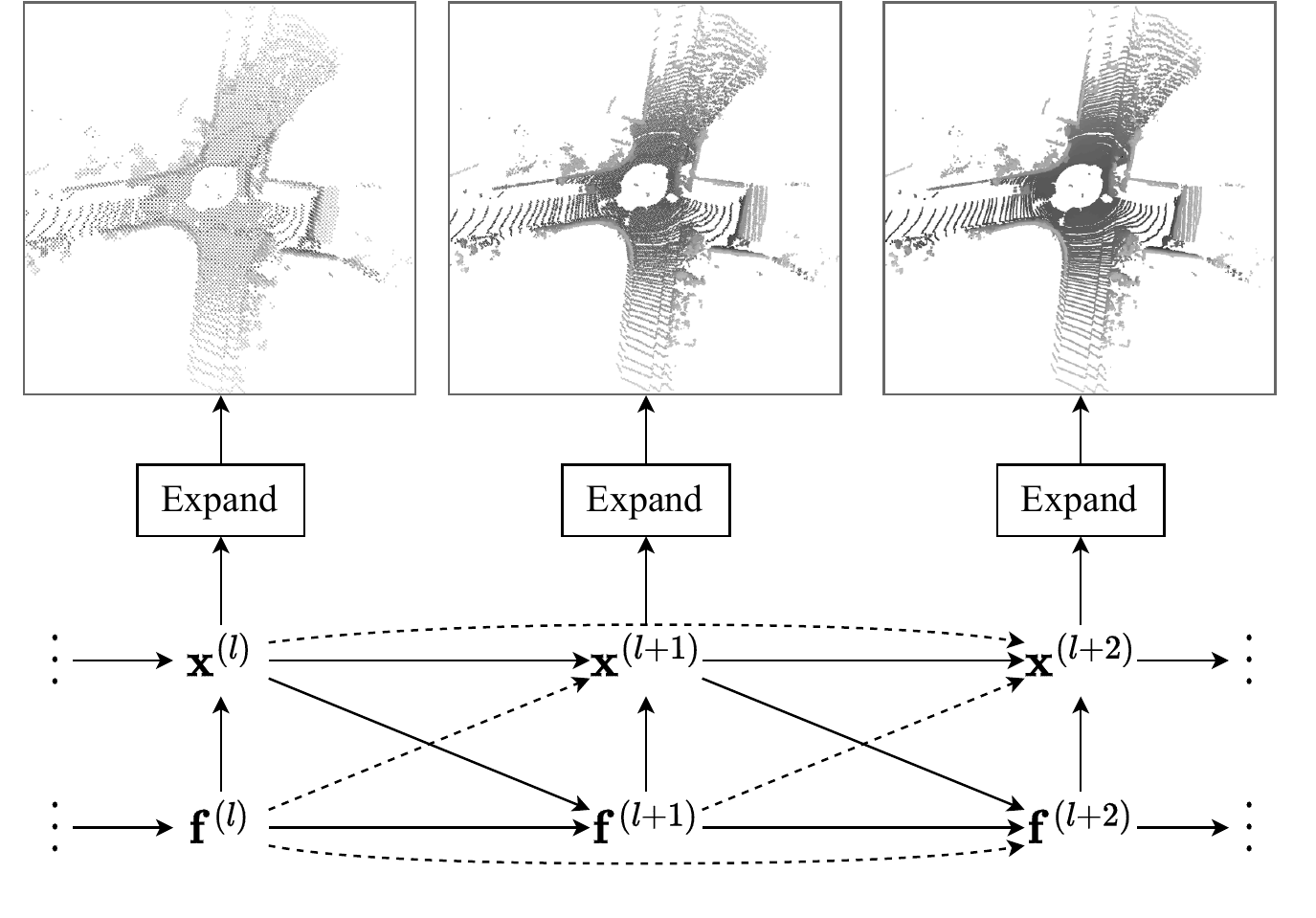}
\caption{The decoding pipeline of the proposed method. The solid and dashed lines represent the dependencies brought by the first- and second-order Markovian properties. The decoding follows the order of $\cdots\rightarrow {\bf f}^{(l)}\rightarrow {\bf x}^{(l)} \rightarrow {\bf f}^{(l+1)} \rightarrow {\bf x}^{(l+1)} \rightarrow \cdots$. The "Expand" denotes the operation that expands the 8-dimensional occupancy map into the corresponding point cloud.}
\label{pic_decoding}
\end{figure}

\section{Problem Formulation}
The proposed framework encodes and decodes the octree in layers to ensure parallelism. The octree is alternatively expressed as an ordered set of layers: ${\bf x}=\{{\bf x}^{(1)},{\bf x}^{(2)},\cdots,{\bf x}^{(L)}\}$, where the decoding of the $l$-th layer ${\bf x}^{(l)}$ depends on the previously decoded layers, i.e., ${\bf x}^{(1:l-1)}$. The corresponding hierarchical probability model is expressed as follows:
\begin{equation}
    q({\bf x})=\prod_l^L q({\bf x}^{(l)}|{\bf x}^{(1:l-1)}).
\end{equation}
The high dimensionality of ${\bf x}^{(l)}$ makes its probability distribution intractable to compute. Instead of relying on the inefficient auto-regressive models for the node-wise decomposition of ${\bf x}^{(l)}$, we introduce a latent variable ${\bf f}^{(l)}$, conditioned on which the nodes in ${\bf x}^{(l)}$ are assumed to be independent \cite{bishop1998latent}. Figure \ref{pic_decoding} shows the decoding pipeline with the introduction of ${\bf f}^{(l)}$. Through ${\bf f}^{(l)}$, the spatial correlation of ${\bf x}^{(l)}$ is captured as follows:
\begin{equation}
    q({\bf x}^{(l)}|{\bf f}^{(l)},{\bf x}^{(1:l-1)})=\prod_i^{N_l}q(x^{(l)}_i|{\bf f}^{(l)},{\bf x}^{(1:l-1)}),
\end{equation}
where $N_l$ is the number of nodes in the $l$-th layer. ${\bf f}^{(l)}$ serves as side information to signal modifications to the probability distribution ${\bf x}^{(l)}$ that requires extra bits to express. Since ${\bf f}^{(l)}$ is to be encoded and decoded together with ${\bf x}^{(l)}$, we express the the joint probability distribution of ${\bf x}^{(l)}$ and ${\bf f}^{(l)}$:
\begin{equation}
    \begin{split}
        q({\bf x}^{(l)},{\bf f}^{(l)}|{\bf x}^{(1:l-1)},{\bf f}^{(1:l-1)};\bm{\psi})=q({\bf x}^{(l)}|{\bf x}^{(1:l-1)},&
        \\
        {\bf f}^{(1:l)};\bm{\psi}_x)q({\bf f}^{(l)}|{\bf x}^{(1:l-1)},{\bf f}^{(1:l-1)};\bm{\psi}_f),&
    \end{split}
\end{equation}
where $\bm{\psi}_x$ and $\bm{\psi}_f$ are the parameters for decoding ${\bf x}^{(l)}$ and ${\bf f}^{(l)}$. By assuming the Markovian property, the joint distribution is reduced to the product of the following two terms:
\begin{equation}\label{equation_xandf}
    \begin{split}
        &q_x^{(l)}=q({\bf x}^{(l)}|{\bf x}^{(l-1)},{\bf f}^{(l)},{\bf x}^{(l-2)},{\bf f}^{(l-1)};\bm{\psi}_x)
        \\
        &q_f^{(l)}=q({\bf f}^{(l)}|{\bf x}^{(l-1)},{\bf f}^{(l-1)},{\bf x}^{(l-2)},{\bf f}^{(l-2)};\bm{\psi}_f).
    \end{split}
\end{equation}
Note that we assume second-order Markovian property to exploit the context of lower layers. $q_x^{(l)}$ can be derived by a classification network with 256 dimensions, each indicating the probability of an occupancy code. However, $q_f^{(l)}$ is unable to be derived in the same way, because different from ${\bf x}^{(l)}$ in one-hot representation, ${\bf f}^{(l)}$ is a numerical vector whose distribution cannot be explicitly represented by a classification network. Therefore, we first apply residual coding to remove the redundancy in ${\bf f}^{(l)}$:
\begin{equation}\label{equation_residual}
\begin{split}
    {\bf f}^{(l)}=\bar{{\bf f}}^{(l)}+{\bf r}^{(l)},{\rm \ with\ }\bar{{\bf f}}^{(l)}=g_d({\bf x}^{(l-1)},{\bf f}^{(l-1)},\\
    {\bf x}^{(l-2)},{\bf f}^{(l-2)};\bm{\psi}_f),
\end{split}
\end{equation}
where $g_d$ is a neural network with $\bm{\psi}_f$ as its parameters, $\bar{{\bf f}}^{(l)}$ is the predictable component given its context, which requires no extra bits; and ${\bf r}^{(l)}$ is the residual. As we have no prior beliefs about ${\bf r}^{(l)}$, we assume it to be independent and identically distributed (i.i.d.), and model it with a fully-factorized entropy model \cite{balle2017end}:
\begin{equation}\label{equation_factorized}
    q_f^{(l)}=q({\bf r}^{(l)})=\prod_i^N q(r^{(l)}_i).
\end{equation}

\begin{figure}[t]
\centering  
\includegraphics[width=0.495\textwidth]{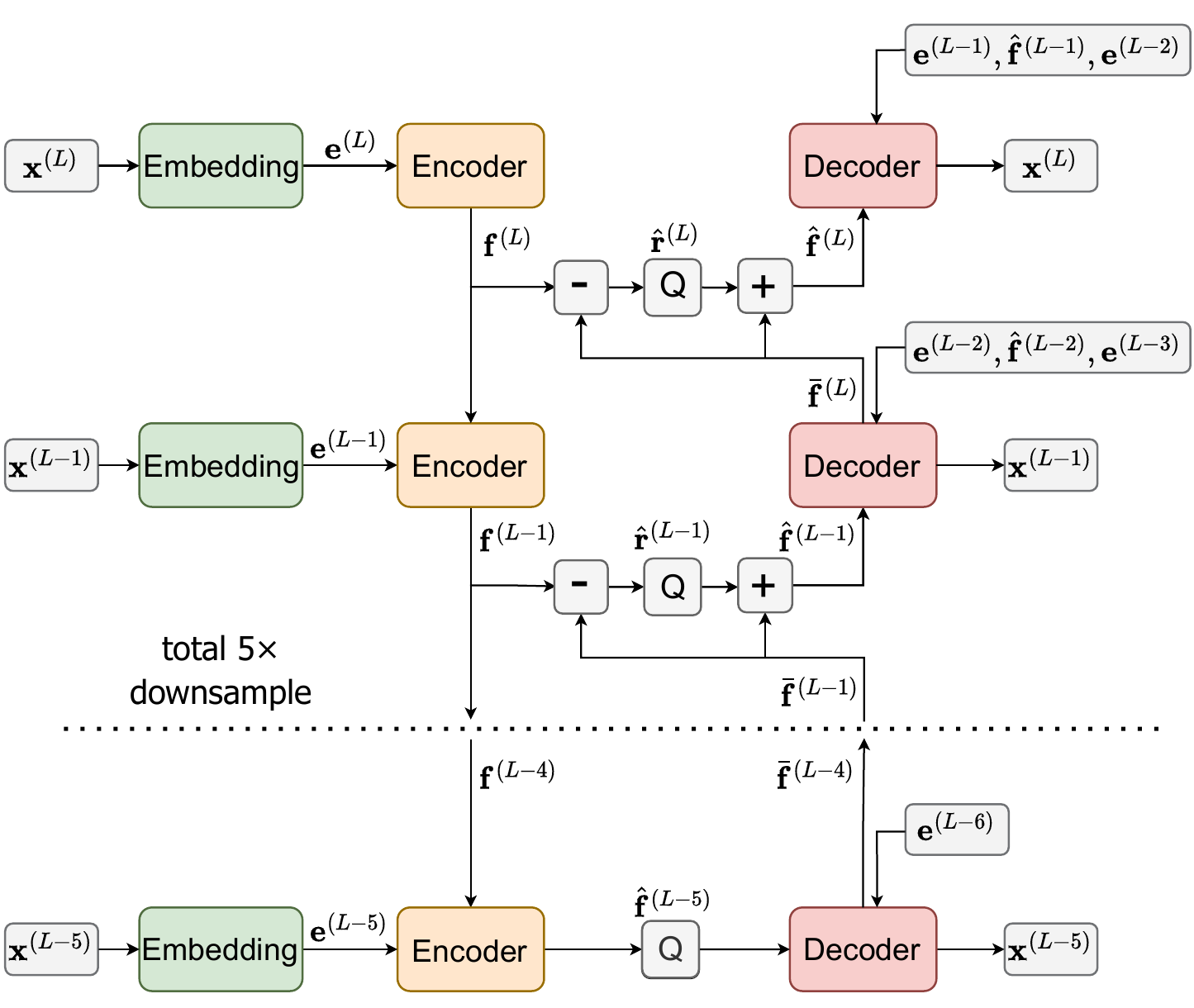}

\caption{The overall architecture of the proposed model. The model consists of 5 layers. For each layer, ${\bf x}^{(l)}$ is the $l$-th octree layer and ${\bf e}^{(l)}$ is the corresponding occupancy embedding. ${\bf f}^{(l)}$ is the latent variable of the $l$-th layer. $\hat{\bf r}^{(l)}$ is the reconstruction residual. $\bar{\bf f}^{(l)}$ and $\hat{\bf f}^{(l)}$ are respectively the predicted and reconstructed latent variable. $\bf -$ and $\bf +$ denote the soft subtraction and soft addition operator. {\bf Q} denotes quantization.}
\label{pic_overall}
\end{figure}

\begin{figure}[t]
\centering  
\includegraphics[width=0.48\textwidth]{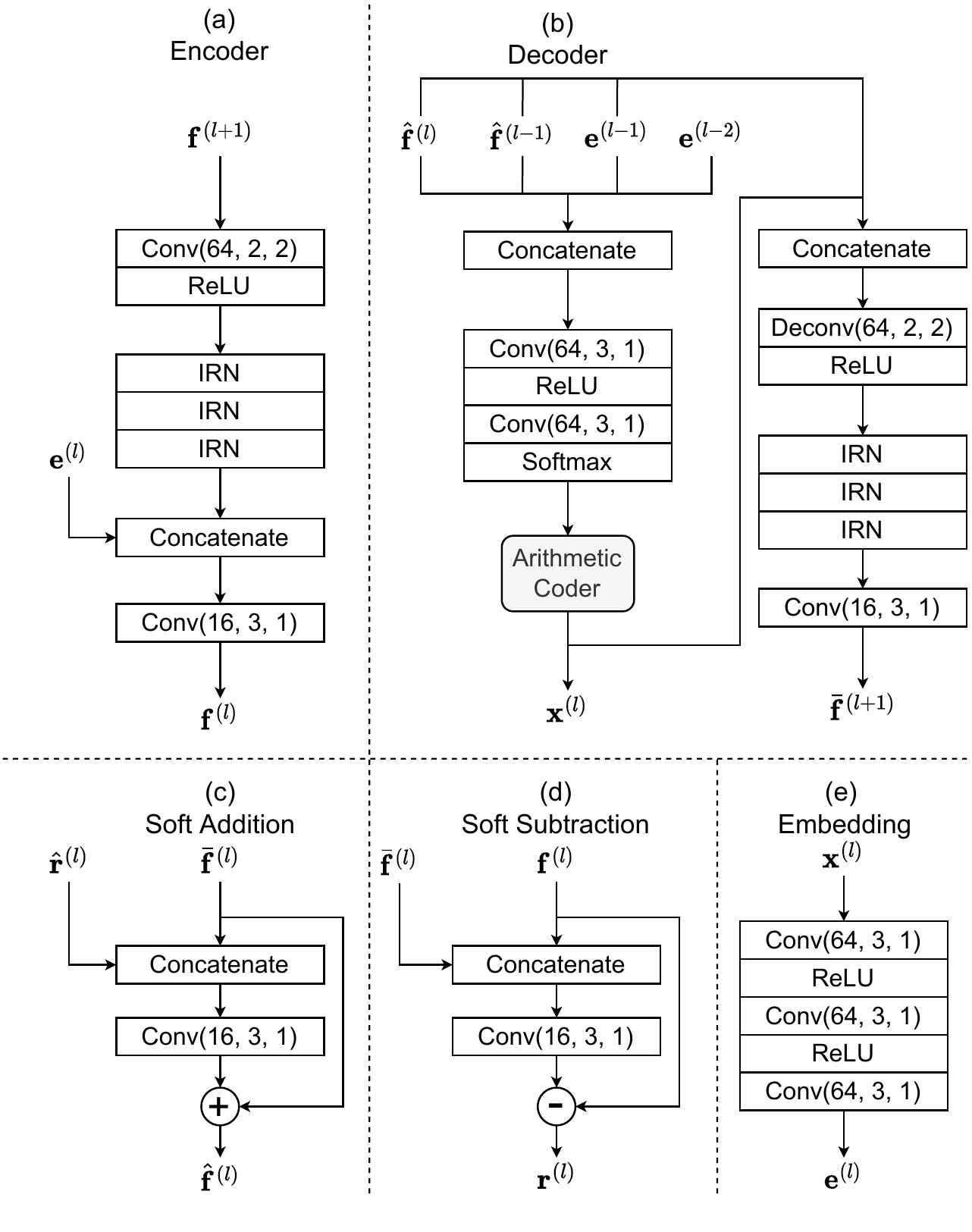}
\caption{The detailed network architecture of (a) the encoder module, (b) the decoder module, (c) the soft addition module, (d) the soft subtraction module, (e) the occupancy embedding module. Conv($C$, $K$, $S$) denotes a sparse convolution layer with output channel size $C$, kernel size $K$, stride $S$. IRN stands for the Inception Residual Network in \cite{wang2021multiscale}.}
\label{pic_enc_dec}
\end{figure}

\section{Network Architecture}
The overall architecture is shown in Figure \ref{pic_overall}. The proposed network follows a 5-layer architecture, where each layer consists of an embedding network, an encoder network, and a decoder network. Each octree layer ${\bf x}^{(l)}$ passes through the embedding network (chapter \ref{chapter_embedding}) to generate the occupancy embedding ${\bf e}^{(l)}$. The encoder network (chapter \ref{chapter_encoder}) analyses the correlation between ${\bf e}^{(l)}$ and ${\bf f}^{(l+1)}$ (assume ${\bf f}^{(L+1)}={\bf 0}$ for consistency) and outputs the latent variable ${\bf f}^{(l)}$ encapsulating the spatial correlation of input variables. The decoder network (chapter \ref{chapter_decoder}) decodes the $l$-th octree layer ${\bf x}^{(l)}$ from the previously decoded context, including $\hat{{\bf f}}^{(l)},{\bf e}^{(l-1)}$, $\hat{{\bf f}}^{(l-1)}$ and ${\bf e}^{(l-2)}$, where $\hat{{\bf f}}^{(i)}$ is the reconstructed latent variable of the $i$-th layer. Meanwhile, the decoder network generates the predicted latent variable $\bar{{\bf f}}^{(l+1)}$ 
for the higher layer, which is {\it softly subtracted} (chapter \ref{chapter_soft}) from ${\bf f}^{(l+1)}$ to generate the residual ${\bf r}^{(l+1)}$. ${\bf r}^{(l+1)}$ is quantized into $\hat{\bf r}^{(l+1)}$ with the quantization step of 1 before entropy coding. During training, the quantization is replaced by adding a uniformly distributed noise $\mu \sim \mathcal{U}(-\frac{1}{2},\frac{1}{2})$ to maintain the differentiablity of the network. Due to the i.i.d. assumption illustrated in Equation (\ref{equation_factorized}), we utilize a factorized entropy model \cite{balle2017end} (chapter \ref{chapter_factorized}) to compress the quantized residual $\hat{{\bf r}}^{(l+1)}$, which is {\it softly added} (chapter \ref{chapter_soft}) with $\bar{{\bf f}}^{(l+1)}$ to generate the reconstructed latent variable $\hat{{\bf f}}^{(l+1)}$ for the higher layer. 

Due to the negligible spatial correlation after down-sampling, $\hat{{\bf f}}^{(L-5)}$ is compressed by a factorized entropy model, and ${\bf x}^{(L-6)}$ is compressed by OctSqueeze \cite{huang2020octsqueeze}. 

\subsection{Occupancy Embedding}\label{chapter_embedding}
The architecture of the occupancy embedding module is shown in Figure \ref{pic_enc_dec}(e). The occupancy embedding module takes the one-hot octree layer ${\bf x}^{(l)}$ as input and outputs the corresponding occupancy embedding ${\bf e}^{(l)}$. The network aims to encode the semantic information of ${\bf x}^{(l)}$ into a densely distributed representation that can be better processed by a neural network. As shown in Figure \ref{pic_enc_dec}(e), the network consists of three sparse convolution layers, which are trained jointly with the entire network. Ablation studies in chapter \ref{chapter_abl_embedding} demonstrate the effectiveness of the occupancy embedding network, where we can observe that occupancy codes geometrically closer are more similar in the vector space.

\subsection{Encoder Network}\label{chapter_encoder}
\begin{figure}[t]
\centering  
\includegraphics[width=0.38\textwidth]{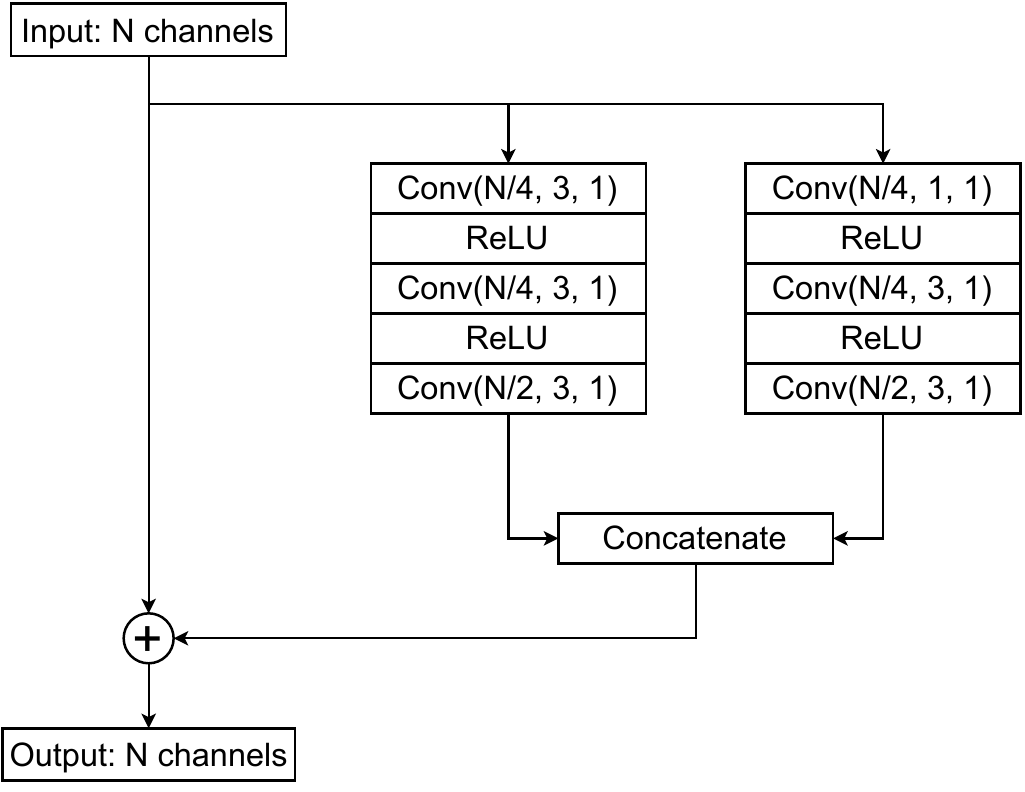}
\caption{The Inception Residual Network (IRN) block.}
\label{pic_incep}
\end{figure}

The architecture of the encoder network is shown in Figure \ref{pic_enc_dec}(a). Equation (\ref{equation_xandf}) shows that the probability distributions of ${\bf x}^{(l)}$ and ${\bf f}^{(l+1)}$ is correlated with latent variable ${\bf f}^{(l)}$ of the $l$-th layer. Therefore, the encoder network (Figure \ref{pic_enc_dec}(a)) is designed as a deterministic transform on ${\bf x}^{(l)}$ and ${\bf f}^{(l+1)}$ to encapsulate their spatial correlation into ${\bf f}^{(l)}$:
\begin{equation}
    {\bf f}^{(l)}=g_e({\bf x}^{(l)},{\bf f}^{(l+1)};\bm{\psi}_e),
\end{equation}
where $\bm{\psi}_e$ is the encoder parameters.
Inspired by Wang {\it et al.} \cite{wang2021multiscale}, we adopt sparse-CNN to design $g_e(\cdot)$ for high-efficiency octree feature extraction. The octree sequence is modeled as a sparse tensor mentioned in equation (\ref{equation_sparse_tensor}), where the coordinate matrix $\bf C$ corresponds to the coordinates of the octree nodes, and the feature matrix $\bf F$ corresponds to the occupancy code. The down-scaling of the octree layers is achieved by a stride-two and kernel-size-two sparse convolution layer. Specifically, ${\bf f}^{(l+1)}$ is first down-sampled by a stride-two sparse convolution layer, followed by several Inception Residual Network (IRN) blocks \cite{szegedy2017inception} (Figure \ref{pic_incep}) for neighborhood feature extraction and aggregation. For the feature fusion of ${\bf f}^{(l+1)}$ and ${\bf x}^{(l)}$, the down-sampled ${\bf f}^{(l+1)}$ is concatenated with the occupancy embedding ${\bf e}^{(l)}$ and passes through a sparse convolution layer to generate the latent variable ${\bf f}^{(l)}$ for the $l$-th layer. 

\subsection{Soft Subtraction/Addition}\label{chapter_soft}
The quantization of the residual ${\bf r}^{(l)}$ introduces error to decoding, leading to noise between the latent variable ${\bf f}^{(l)}$ and its reconstruction $\hat{{\bf f}}^{(l)}$. Therefore, instead of directly adding the quantized residual $\hat{{\bf r}}^{(l)}$ to the predicted latent variable $\bar{{\bf f}}^{(l)}$ shown in equation (\ref{equation_residual}), we propose {\it soft addition} (denoted as $\oplus$) to compensate for the information loss during quantization:
\begin{equation}\label{equation_soft_addition}
    \hat{{\bf f}}^{(l)}=\bar{{\bf f}}^{(l)} \oplus \hat{{\bf r}}^{(l)}=\bar{{\bf f}}^{(l)}+h_a(\hat{{\bf r}}^{(l)},\bar{{\bf f}}^{(l)};\bm{\psi}_a),
\end{equation}
where $h_a(\cdot)$ is a transform with $\bm{\psi}_a$ as its parameters. Similarly, the {\it soft subtraction} (denoted as $\ominus$) is defined as follows:
\begin{equation}\label{equation_soft_subtraction}
   {\bf r}^{(l)}={\bf f}^{(l)} \ominus \bar{{\bf f}}^{(l)}={\bf f}^{(l)}-h_s(\bar{{\bf f}}^{(l)},{\bf f}^{(l)};\bm{\psi}_s).
\end{equation}
Note that the difference between $\bar{{\bf f}}^{(l)}$ and ${\bf f}^{(l)}$ can be arbitrarily close to zero provided that with additional bit rate to encode $\hat{{\bf r}}^{(l)}$ and the corresponding proper choice of $h_a(\cdot)$, $h_s(\cdot)$. The network dynamically adjusts the quantization precision of each node in $\hat{{\bf f}}^{(l)}$ to balance the reconstruction quality of $\hat{{\bf f}}^{(l)}$ and the corresponding bit rate consumption in an end-to-end manner. $h_a(\cdot)$ and $h_s(\cdot)$ (Figure \ref{pic_enc_dec}(c)(d)) are designed as sparse convolution layers to fit arbitrary transformations.

\subsection{Decoder Network}\label{chapter_decoder}
The design of the decoder network follows equation (\ref{equation_xandf}) and (\ref{equation_residual}). When constructing the network, we assume second-order Markovian property for the system stability, as the lower layer context is potentially informative for the entropy coding of the current layer. To demonstrate the necessity of second-order Markov property, we compare the performance of first- and second-order Markov assumptions in chapter \ref{chapter_abl_2level}.

Figure \ref{pic_enc_dec}(b) shows the architecture of the decoder network, which is divided into two branches corresponding to the decoding of ${\bf x}^{(l)}$ and $\bar{{\bf f}}^{(l+1)}$ respectively:
\subsubsection{The decoding of ${\bf x}^{(l)}$}
The context of ${\bf x}^{(l)}$, including ${\bf x}^{(l-1)}$, ${\bf f}^{(l)}$, ${\bf x}^{(l-2)}$ and ${\bf f}^{(l-1)}$ are concatenated together and pass through a 2-layer sparse CNN. Finally, a 256-dimensional softmax layer is used to generate the probability distribution of ${\bf x}^{(l)}$ (denoted as ${\bf p}^{(l)}\in \mathbb{R}^{N_l\times 256}$). During training, we directly represent the bit rate of entropy coding with the cross entropy between ${\bf x}^{(l)}$ and ${\bf p}^{(l)}$, since cross entropy is a tight lower bound achievable using arithmetic coding algorithms. When evaluating the model, an arithmetic coder is used to encode the difference between ${\bf p}^{(l)}$ and ${\bf x}^{(l)}$ to ensure that the decoding of ${\bf x}^{(l)}$ is lossless.
\subsubsection{The decoding of $\bar{{\bf f}}^{(l+1)}$}
The context ${\bf x}^{(l)}$, ${\bf f}^{(l)}$, ${\bf x}^{(l-1)}$ and ${\bf f}^{(l-1)}$ are concatenated together and up-sampled to generate the predicted latent variable $\bar{{\bf f}}^{(l+1)}$. We adopt a stride-two and kernel-size-two sparse transposed convolution layer to achieve the octree up-sampling. Similar to the encoder network, we utilize several IRN modules for local feature analysis and aggregation and a stride-one sparse convolution layer to adjust the dimension of output.

Note that the nodes in the same layer share the same context, meaning that the decoding process of each node is independent of other nodes in the same layer. This independence brings superior parallelism to the entire network. Compared with OctAttention, which requires re-running the model multiple times to overcome the continuous varying of context, our proposed model needs only run {\bf once} during decoding. Therefore, one can observe significant runtime improvement in the subsequent experiments.

\subsection{Factorized Deep Entropy Model}\label{chapter_factorized}
We adopt the factorized deep entropy model \cite{balle2017end} for entropy coding. Two operations in the entropy encoding system block the back-propagation of gradients: 1) quantizing the input variable $\bf y$ into $\hat{\bf y}$, 2) encoding the quantized variable to and from the bitstream. For 1), the quantization operation is replaced by adding a uniformly distributed noise $\bm{\mu} \sim \mathcal{U}(-\frac{1}{2},\frac{1}{2})$ to ensure differentiability during training. 

For 2), thanks to the development of arithmetic coding, we can approximate the bit rate of encoding $\hat{\bf y}$ without the actual encoding and decoding process. The estimated size of bitstream is the entropy of $\hat{\bf y}$, which is a tight lower bound achievable by arithmetic coding algorithms:
\begin{equation}
    \mathcal{R}_{\hat{\bf y}}=\mathbb{E}_{\hat{\bf y}}\left[-\log q_{\hat{\bf y}}(\hat{\bf y})\right]
\end{equation}

We follow \cite{balle2017end} and model the probability distribution of $\hat{\bf y}$ with a fully factorized density model:
\begin{equation}
    q_{\hat{\bf y}|\bm{\phi}}(\hat{\bf y}|\bm{\phi})=\prod_i \left(q_{y_i|\bm{\phi}^{(i)}}(\bm{\phi}^{(i)})*\mathcal{U}(-\frac{1}{2},\frac{1}{2})\right)(\hat{y}_i),
\end{equation}
where $q_{y_i|\bm{\phi}^{(i)}}$ is the univariate distribution of the $i$-th channel and $\bm{\phi}^{(i)}$ is the corresponding parameters. In general, adding a hyperprior variable $\bf z$ upon $\hat{\bf y}$ can further capture the spatial correlation of $\hat{\bf y}$ for more bit rate reduction \cite{balle2018variational,hu2020coarse}. However, it is unnecessary for our proposed model due to the i.i.d. assumption of the residual ${\bf r}^{(l)}$ in equation (\ref{equation_factorized}). More information can be found in Section \ref{chapter_ablation_latent} that plots the visualization of ${\bf r}^{(l)}$. One can observe the strong independence of ${\bf r}^{(l)}$ among nodes in the same layer.
\subsection{Loss function}\label{chapter_loss_function}
The model is trained in an end-to-end manner with the following loss function:
\begin{equation}\label{equation_loss}
    \mathcal{L}=\alpha \sum_{l=L-k}^{L} \mathcal{R}_r^{(l)}+\sum_{l=L-k}^{L}\mathcal{R}_x^{(l)},
\end{equation}
where $k$ is the number of layers, $\mathcal{R}_r^{(l)}$ is the {\it bits per output point} bpop to encode the quantized residual $\hat{\bf r}^{(l)}$ (let $\hat{\bf r}^{(L-k)}$=$\hat{\bf f}^{(L-k)}$ for consistency) of the $l$-th layer, $\mathcal{R}_x^{(l)}$ is the bpop to entropy encode ${\bf x}^{(l)}$. In experiments, we find that the model easily converges to a local optimum where the bit rate of $\hat{\bf f}^{(l)}$ is extremely low. Therefore, we add a scaling factor $\alpha$ to encourage the model to output more informative $\hat{\bf f}^{(l)}$. $\alpha$ is set as $0.5$ for the first several epochs and $0.95$ after that. $\mathcal{R}_r^{(l)}$ is computed by the factorized entropy model mentioned in chapter \ref{chapter_factorized}. $\mathcal{R}_x^{(l)}$ is computed by the binary cross entropy (BCE) between the probability ${\bf p}^{(l)}$ and the ground truth ${\bf x}^{(l)}$:
\begin{equation}
    \mathcal{R}_x^{(l)}=-\frac{1}{N_l}\sum_{i=1}^{N_l}\sum_{j=1}^{256}x_{ij}^{(l)}\log_2 p_{ij}^{(l)},
\end{equation}
where $N_l$ is the number of nodes in the $l$-th layer, and $N$ is the number of points in the output point cloud.

\section{Experiments}
\begin{figure}[t]
\centering  
\includegraphics[width=0.495\textwidth]{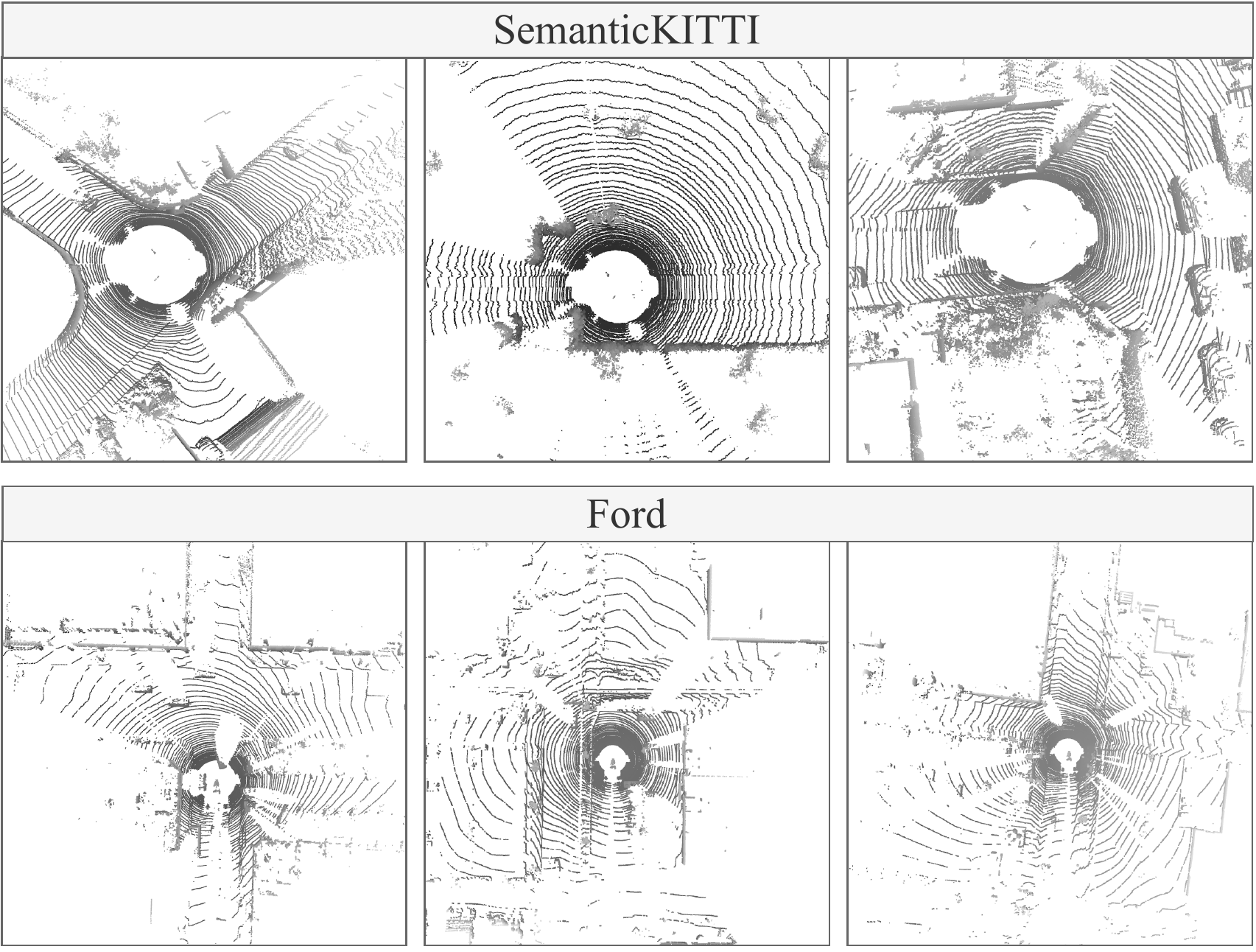}
\caption{The visualization of datasets. We randomly choose six scans from two datasets. The upper row is the scans from SemanticKITTI. The lower row is the scans from Ford.}
\label{pic_dataset}
\end{figure}
\subsection{Datasets}
\begin{itemize}
    \item {\bf SemanticKITTI} \cite{behley2019semantickitti} is a public LiDAR dataset for autonomous driving, containing 22 sequences with 43352 scans collected from a Velodyne HDL-64 sensor. We adopt the official data splitting that takes sequences 00 to 10 as the training set and 11 to 21 as the test set.
    \item {\bf Ford} is a LiDAR dataset suggested by MPEG Common Test Conditions (CTC) \cite{schwarz2018common}. The dataset contains three sequences with 4500 frames. The precision of the original point cloud is 18 bits. We divide sequence 01 as the training set and sequences 02 and 03 as the test set.
\end{itemize}

\begin{table}[t]
     \caption{Different BD-Rate(\%) gains against G-PCC.}
    \centering
    \setlength{\tabcolsep}{1.2mm}{
    \begin{tabular}{cc cc cc cc cc}
      \toprule
      \multicolumn{2}{c}{} & Ours & OctAttention & Voxelcontext-Net & OctSqueeze  \\ 
      \midrule
      \multicolumn{2}{c}{SemanticKITTI} & \textbf{-28.27} & -25.33 & -14.37 & -3.95 \\
      \multicolumn{2}{c}{Ford} & \textbf{-25.55} & -22.01 & -11.68 & -2.98 \\
      \bottomrule
     \end{tabular}
     }
      \label{tab:bd}
\end{table}

\subsection{Training Strategy}
\begin{itemize}
    \item For SemanticKITTI, we train four models for octree depth $L=12,11,10,9$. We utilize an Adam optimizer with $\beta=(0.9,0.999)$ and the initial learning rate set as $0.0006$. We further utilize a learning rate scheduler with a decay rate of $0.7$ for every 20 epochs. We train the proposed model for 200 epochs, with 1200 iterations for each epoch. For the first 10 epochs, $\alpha$ in equation (\ref{equation_loss}) is set as 0.5. The batch size is set as 4 during training. We conduct all the experiments on a GeForce RTX 3090 GPU with 24GB memory.
    \item For Ford, we also train four models for octree depth $L=12,11,10,9$. As Ford is a small dataset with only 1500 frames for training, we first train the $L=12$ model for 40 epochs. $\alpha$ in equation (\ref{equation_loss}) is set as 0.5 for the first two epochs and 0.95 for the remaining. For the models of $L=11,10,9$, since the amount of quantized data is too small to support the end-to-end training, we perform four epochs of finetuning on the $L=12$ model on {\bf Ford} dataset with the initial finetuning learning rate 0.0002. $\alpha$ is set as 0.95 during finetuning.
\end{itemize}

\begin{figure*}[htbp]
\centering  
\includegraphics[width=0.95\textwidth]{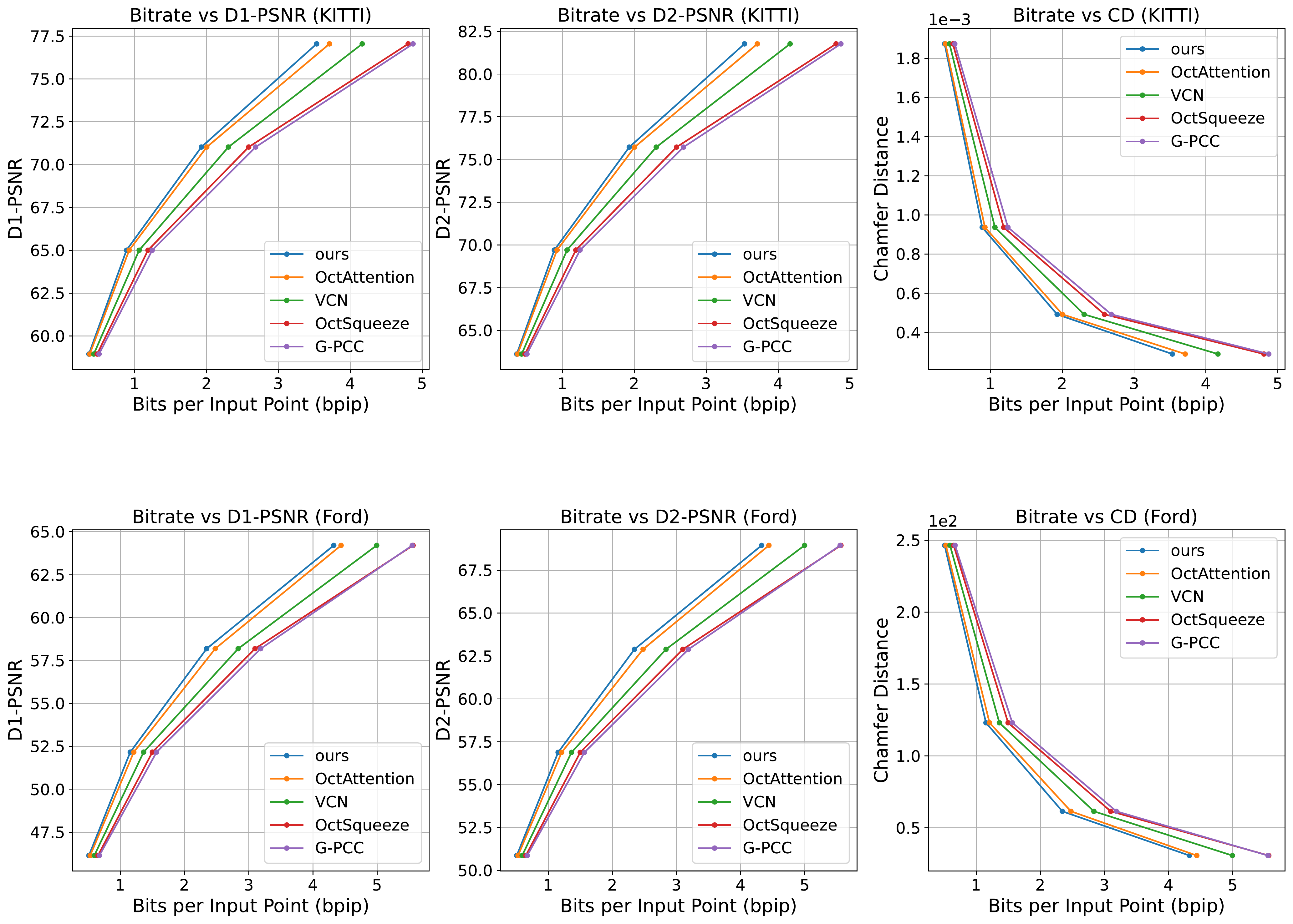}
\caption{Rate-distortion curves on SemanticKITTI and Ford. From left to right: D1-PSNR, D2-PSNR and chamfer distance}
\label{pic_kitti}
\end{figure*}

\begin{figure}[t]
\centering  
\includegraphics[width=0.48\textwidth]{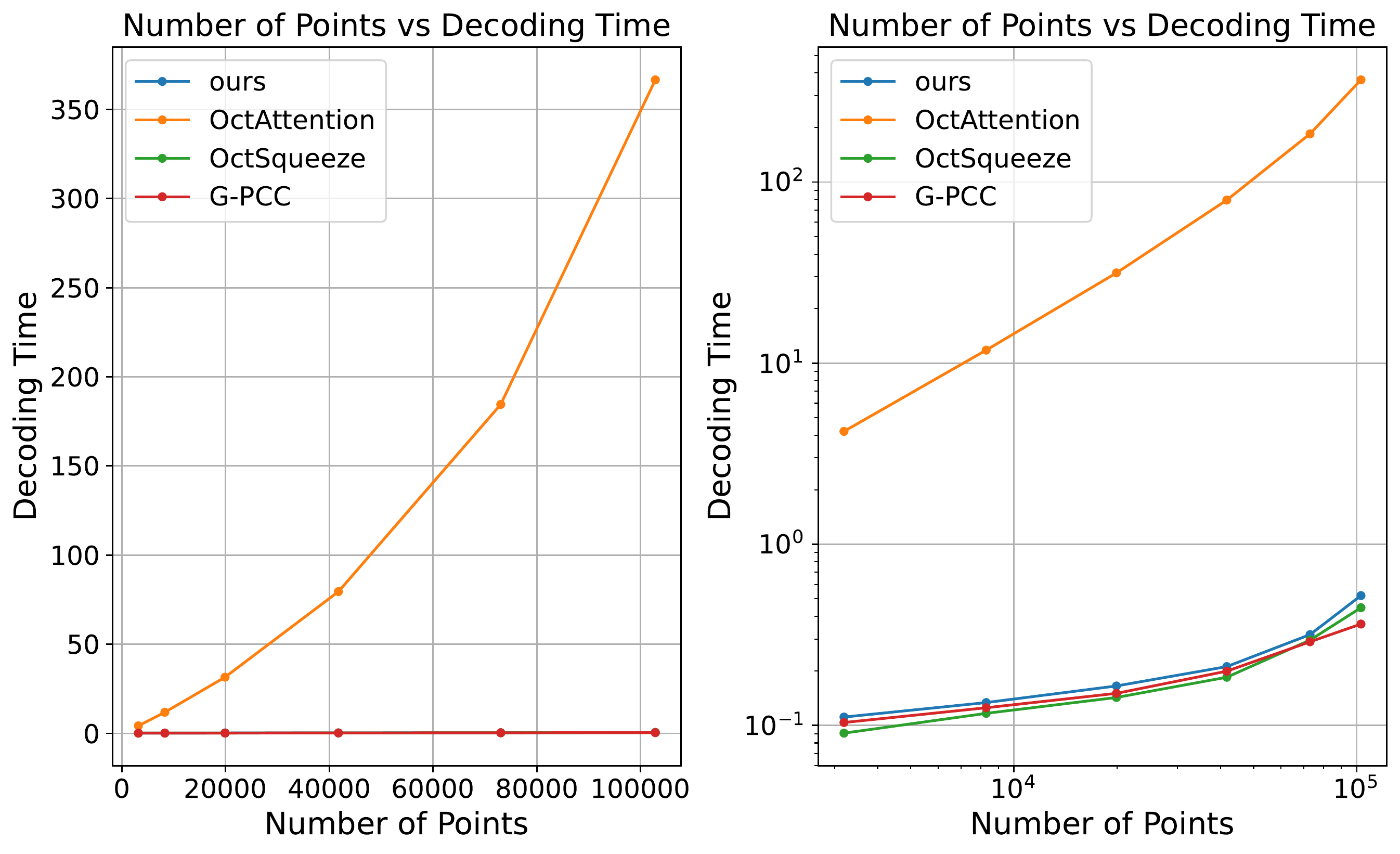}
\caption{Decoding time comparison of our model, OctAttention, OctSqueeze and G-PCC. The experiment is done on SemanticKITTI-00-000000. For each network, we run it ten times and count the average decoding time. The left figure is the decoding time comparison of the ordinary coordinate. The right figure is the experiment results of the logarithmic coordinate.}
\label{pic_runtime}
\end{figure}

\subsection{Baseline Setup}
We compare our proposed model with G-PCC and other learning-based LPC compression network including OctSqueeze, VoxelContext-Net and state-of-the-art method OctAttention:
\begin{itemize}
    \item For G-PCC, we follow the MPEG CTC to generate the results on SemanticKITTI and Ford with the latest TMC13-v14.
    \item For OctSqueeze, as the source code is not publicly available, we re-implement their model and train for 200 epochs on SemanticKITTI, 40 epochs on Ford 01.
    \item For VoxelContext-Net, as the source code is also unavailable, we adopt the re-implemented result by Muhammad et al. \cite{muh2022vox}, which reports better results than \cite{que2021voxelcontext}. We remove the coordinate refinement module (CRM) in VoxelContext-Net for the fairness of comparison. It is a post-processing module that enhances the quantized point cloud to reduce distortion but is irrelevant from the entropy coding of point cloud geometry.
    \item For OctAttention, we adopt the result on SemanticKITTI in the paper. Since they have no available results on the Ford dataset, we retrain their model on Ford 01 for 8 epochs (the same number of epochs as SemanticKITTI) to generate the result on Ford 02 and 03.
\end{itemize}

\begin{figure*}[t]
\centering  
\includegraphics[width=0.95\textwidth]{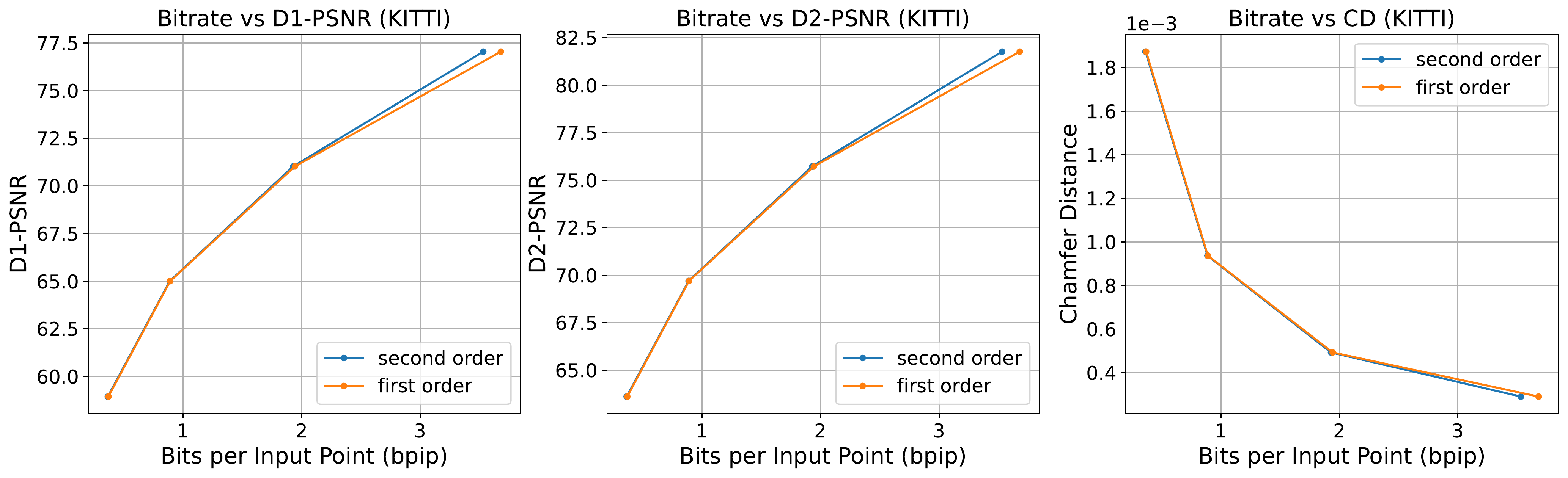}
\caption{Ablation study of second-order Markovian property and first-order Markovian property.}
\label{pic_Markovian}
\end{figure*}

\begin{figure*}[t]
\centering  
\includegraphics[width=0.95\textwidth]{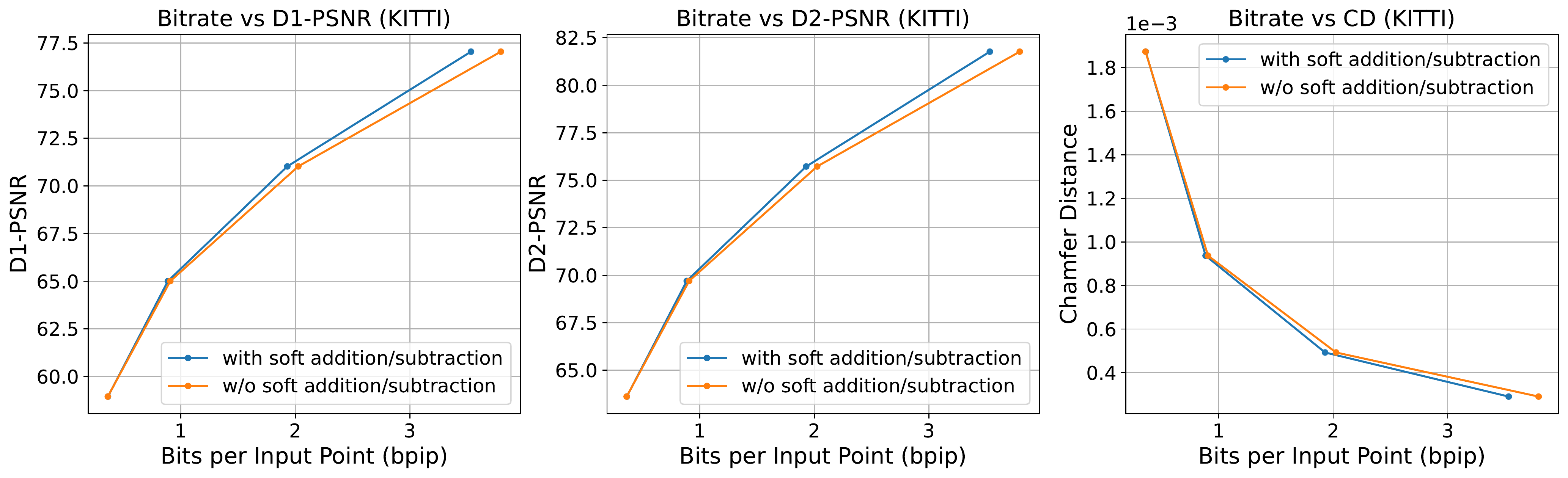}
\caption{Ablation study of the model with soft addition/subtraction and w/o addition/subtraction.}
\label{pic_soft}
\end{figure*}

\subsection{Evaluation Metrics}
The bit rate is evaluated using bits per input point (bpip), and the distortion is evaluated using point-to-point geometry (D1) Peak Signal-to-Noise Ratio (PSNR) and point-to-plane geometry (D2) PSNR, which are computed by the software {\it pc\_error} developed by MPEG. For Ford dataset, we follow the MPEG CTC that sets the PSNR peak value $p=30000$. For the floating-point SemanticKITTI dataset, we follow the evaluation metric in \cite{fu2022octattention} that scales the point clouds into $[-1,1]^3$, and sets the PSNR peak value $p=1$. We also compare the chamfer distance (CD) among all methods:
\begin{equation}
\begin{split}
    &{\rm CD}({\bf P},\bar{\bf P})={\rm max}({\rm CD}^{'}({\bf P},\bar{\bf P}),{\rm CD}^{'}(\bar{\bf P},{\bf P})),\\
    &{\rm CD}^{'}({\bf P},\bar{\bf P})=\frac{1}{|{\bf P}|}\sum_i {\rm min}_j \left \| {\bf p}_i-\bar{\bf p}_j \right \|_2.
\end{split}
\end{equation}

\subsection{Experiment Results}
\subsubsection{Results on SemanticKITTI}
The rate-distortion (D1-PSNR, D2-PSNR, chamfer distance) curves of different methods on SemanticKITTI are shown in Figure \ref{pic_kitti}. We can observe that the four learning-based methods outperform G-PCC due to the optimization by traversing the whole dataset. Besides, OctSqueeze and Voxelcontext-Net assume different degrees of node independence considering the time complexity; thus, neither fully explores the sibling correlation, leading to suboptimal performance. Our method dramatically surpasses the two methods due to the hierarchical latent variable, which encapsulates the correlation of sibling nodes to remove spatial redundancy. Meanwhile, our model surpasses OctAttention, which considers sibling dependence despite the excessive complexity. We also report the corresponding BD-Rate gain of the four learning-based methods against G-PCC in Table \ref{tab:bd}. The experiment result shows that our model achieves state-of-the-art coding efficiency on SemanticKITTI, with the highest -28.27\% BD-Rate gain against G-PCC.
\subsubsection{Results on Ford}
The rate-distortion curves of different methods on Ford are also shown in Figure \ref{pic_kitti}, and the corresponding BD-rate gain against G-PCC is shown in Table \ref{tab:bd}. Ford is a dataset suggested by MPEG CTC. Our proposed model also achieves state-of-the-art performance on Ford, with the highest -25.55\% BD-Rate gain against G-PCC. The experiment result demonstrates the extensibility of our model.
\subsubsection{Complexity Comparison}

\begin{figure*}[htbp]
\centering  
\includegraphics[width=0.95\textwidth]{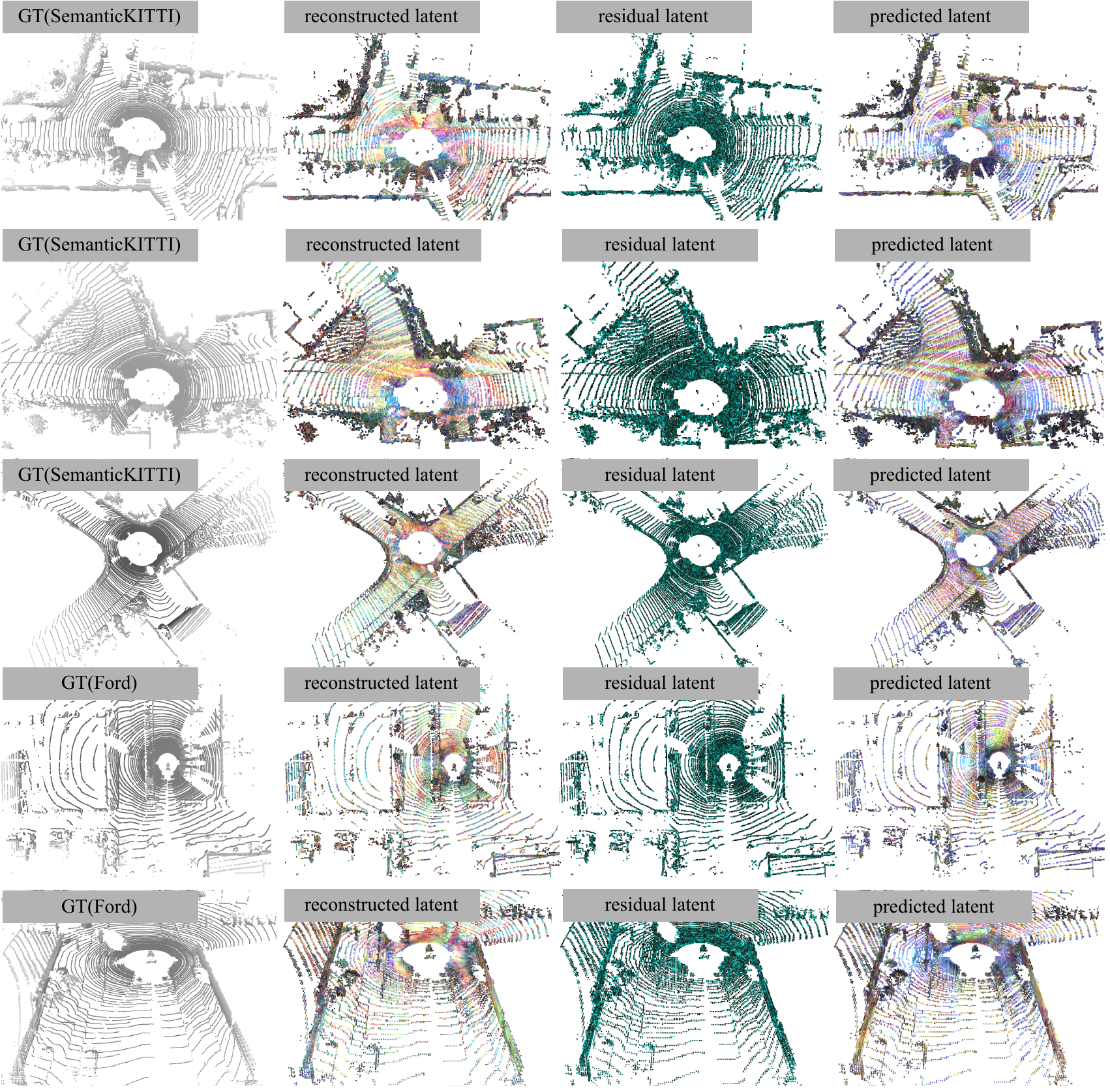}
\caption{The visualization of the latent variable and residual. We randomly visualize 5 LiDAR point clouds and the corresponding latent variables and residuals from SemanticKITTI and Ford. The 5 point clouds are from separate sequences. The leftmost column shows the original LiDAR point cloud. The middle-left column shows the reconstructed latent variable $\hat{\bf f}^{(l)}$. The middle-right column shows the residual $\hat{\bf r}^{(l)}$. The rightmost column shows the predicted component of $\hat{\bf f}^{(l)}$, i.e. $\bar{\bf f}^{(l)}$. The color is visualized by random projection \cite{bingham2001random}, which projects the high dimensional latent variable and residual into 3D color space, then normalizes the 3D vectors to 0-255 to represent the RGB value. The dependence can be judged by color. The area with similar colors can be regarded as highly spatial-correlated.}
\label{pic_residual}
\end{figure*}

In this section, we evaluate the complexity. Figure \ref{pic_runtime} compares the decoding time of each method. We can observe that:
\begin{itemize}
    \item The complexity of OctAttention is extraordinarily high compared to other methods. This is due to the sibling context selection strategy, which causes the context to change during the decoding process and requires multiple runs of the model to update the context. OctAttention requires more than 5 minutes to decode one point cloud frame, which is unaffordable in practical applications.
    \item On the contrary, our model is end-to-end and fully-factorized, where the nodes in the same layer are independently decoded. Compared with OctAttention, our model is highly parallelized and requires running only {\bf once}, thus saving more than 99.8\% of decoding time. 
    \item Due to the complexer network structure, our model is slightly slower than OctSqueeze, which assumes the independence of sibling nodes and also decodes layer-wisely. But our model explores the correlation of sibling nodes, thus our model achieves better performance.
\end{itemize}

\section{Analysis and Ablation Study}


\subsection{Ablation study on soft addition/subtraction}
In chapter \ref{chapter_soft}, we propose substituting the ordinary addition/subtraction operator for the soft addition/subtraction. Therefore, we perform an ablation study on its effectiveness in Figure \ref{pic_soft}. We separately train the two models on the training split of SemanticKITTI and evaluate the result on the test split. The experiment result proves that the soft addition/subtraction enables the network to dynamically balance the reconstruction loss and the corresponding bit rate of $\hat{\bf f}^{(l)}$, thus achieving 3.47\% BD-Rate gain against the ordinary addition/subtraction.

\subsection{Ablation study on the Markovian property}\label{chapter_abl_2level}
In equation (\ref{equation_xandf}), we assume the Markovian property of ${\bf f}^{(l)}$ and ${\bf x}^{(l)}$ for simplification. When establishing the network, we instead assume second-order Markovian property for system stability in the hope that the lower layer context is also informative for the entropy coding of the current layer. We separately train two models of first-order and second-order Markovian properties on the training split of SemanticKITTI and evaluate the result on the test split. In Figure \ref{pic_Markovian}, we show the performance comparison of the model based on first-order and second-order Markovian properties. The experimental result proves that the assumption of second-order Markovian property is effective, which achieves 1.09\% BD-Rate gain against first-order Markovian property.

\subsection{Analysis on Latent Variable}\label{chapter_ablation_latent}

\begin{figure}[tbp]
\centering  
\includegraphics[width=0.45\textwidth]{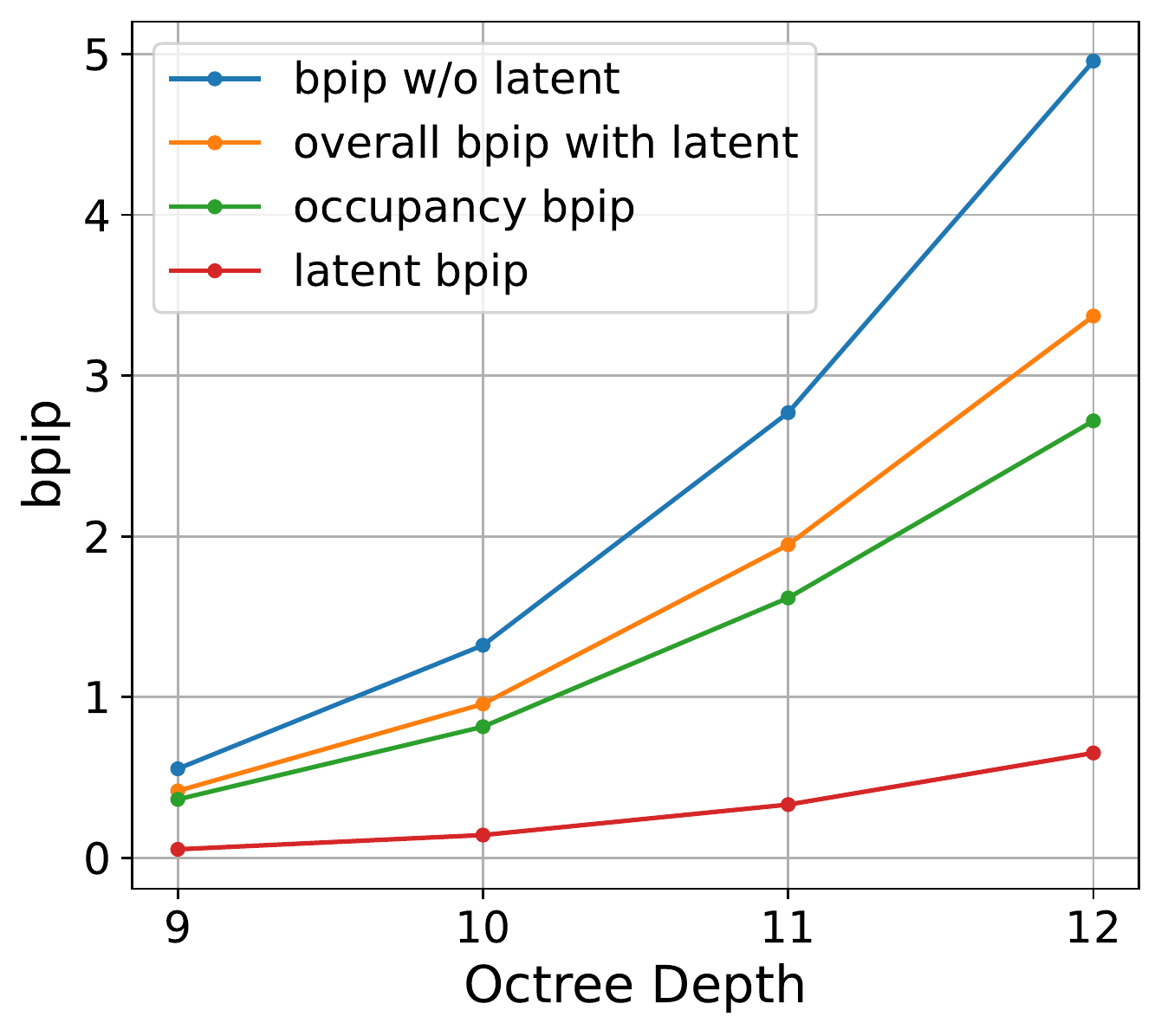}
\caption{The overall bit rate of the entropy model with latent variable (orange line), the bit rate of the occupancy with latent variable (green line), the bit rate of the latent variable (red line), and the overall bit rate of the entropy model if no latent variable is introduced to capture the sibling dependency (blue line). The x-axis is the depth of the octree, i.e., $L$. The performance comparison is made on SemanticKITTI-00-000000.}
\label{pic_latent}
\end{figure}

Defined by equation (\ref{equation_residual}), the latent variable $\hat{\bf f}^{(l)}$ of the $l$-th octree layer is decomposed into two parts: the predictable component $\bar{\bf f}^{(l)}$ given the previously decoded context, and the independent residual component $\hat{\bf r}^{(l)}$ with no prior knowledge. In this section, we visualize the decomposition process of $\hat{\bf f}^{(l)}$ in Figure \ref{pic_residual}. The middle-left column is the visualization of $\hat{\bf f}^{(l)}$, with clearly visible spatial dependence among each node. The middle-right column is the visualization of the residual $\hat{\bf r}^{(l)}$ that captures the unpredictable component of $\hat{\bf f}^{(l)}$. We can observe that $\hat{\bf r}^{(l)}$ consists of a large number of noisy points, which demonstrates the i.i.d. 
nature of $\hat{\bf r}^{(l)}$ in equation (\ref{equation_factorized}). The rightmost column is the visualization of predictable component $\bar{\bf f}^{(l)}$ that captures the sibling dependence of $\hat{\bf f}^{(l)}$, thus we can observe stronger spatial correlation than $\hat{\bf f}^{(l)}$. The visualization indicates that the residual coding successfully captures the dependence of high-scale data into a low-scale variable, and guarantees the i.i.d. nature of the remaining information.


Figure \ref{pic_latent} shows the bit rate consumption to encode the latent variable and occupancy, and the effect of the latent variable on encoding occupancy. The bit rate of the latent variable increases slowly with the overall bit rate, accounting for less than 10\%. However, the benefit of introducing latent variables as side information is significant. The bit rate of encoding the occupancy is reduced by more than 40\% due to the latent variable. Note that the network autonomously balances the bit rate of each part via end-to-end training, with the final bit rate as the optimization goal.

\begin{figure}[t]
\centering  
\includegraphics[width=0.47\textwidth]{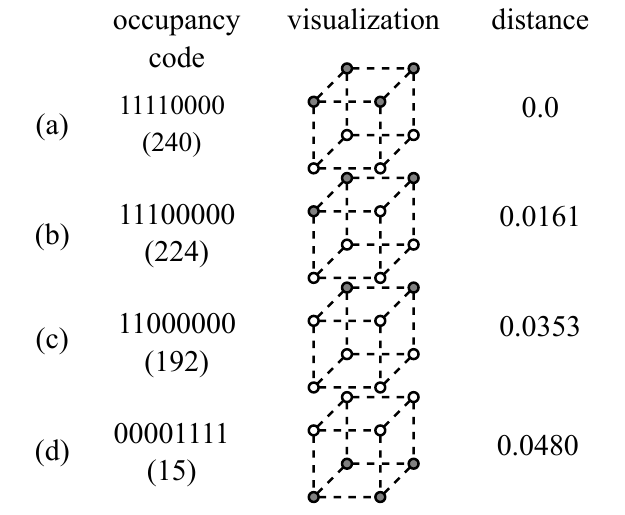}
\caption{Analysis on occupancy embedding module. The left column is the occupancy code of octree nodes. The middle column is the visualization of the octree node. The right column is the Euclidean distance of the feature of octree nodes from the octree node $11110000$ ($240$) in feature space.}
\label{pic_emb}
\end{figure}
\subsection{Analysis on Occupancy Embedding Module}\label{chapter_abl_embedding}
To validate the effectiveness of the occupancy embedding module, we compute the feature embedding of different occupancy codes. In Figure \ref{pic_emb}, we compare the Euclidean distance of three occupancy codes from the base occupancy code $11110000$ ($240$). Note that the input occupancy codes are one-hot vectors whose distance is one between each other. However, after being converted by the embedding module, occupancy codes that are geometrically closer are similar in the feature space, demonstrating that the occupancy embedding module learns the geometric semantics of the occupancy codes.
\section{Conclusion}
This paper proposes an end-to-end, fully-factorized LiDAR point cloud compression framework based on the hierarchical decomposition of the octree entropy model. The network utilizes a latent variable to encapsulate the sibling and ancestor dependence, through which the sibling nodes are conditionally independent to be decoded in parallel. We further propose a residual coding framework for the entropy coding of the latent variable. The latent variable is hierarchically downsampled to capture the deeper spatial correlation, and the residual is modeled by a factorized entropy model. We propose to use soft addition/subtraction to substitute ordinary addition/subtraction for the flexibility of the network. The experiment result shows that our framework achieves state-of-the-art performance among all the LPC compression frameworks. It is noted that our framework also shows superior time complexity, saving more than 99.8\% decoding time compared with the previous state-of-the-art framework.



%



\section*{Acknowledgment}
This paper is supported in part by National Natural Science Foundation of China (61971282, U20A20185). The corresponding author is Yiling Xu(e-mail: yl.xu@sjtu.edu.cn).

\ifCLASSOPTIONcaptionsoff
  \newpage
\fi



%


\bibliographystyle{IEEEtran}
\bibliography{b}

\begin{thebibliography}{10}
\providecommand{\url}[1]{#1}
\csname url@samestyle\endcsname
\providecommand{\newblock}{\relax}
\providecommand{\bibinfo}[2]{#2}
\providecommand{\BIBentrySTDinterwordspacing}{\spaceskip=0pt\relax}
\providecommand{\BIBentryALTinterwordstretchfactor}{4}
\providecommand{\BIBentryALTinterwordspacing}{\spaceskip=\fontdimen2\font plus
\BIBentryALTinterwordstretchfactor\fontdimen3\font minus
  \fontdimen4\font\relax}
\providecommand{\BIBforeignlanguage}[2]{{%
\expandafter\ifx\csname l@#1\endcsname\relax
\typeout{** WARNING: IEEEtran.bst: No hyphenation pattern has been}%
\typeout{** loaded for the language `#1'. Using the pattern for}%
\typeout{** the default language instead.}%
\else
\language=\csname l@#1\endcsname
\fi
#2}}
\providecommand{\BIBdecl}{\relax}
\BIBdecl

\bibitem{xu2018introduction}
Y.~Xu, K.~Zhang, L.~He, Z.~Jiang, and W.~Zhu, ``Introduction to point cloud
  compression,'' \emph{ZTE Communications}, vol.~16, no.~3, p.~8, 2018.

\bibitem{zhu2020view}
W.~Zhu, Z.~Ma, Y.~Xu, L.~Li, and Z.~Li, ``View-dependent dynamic point cloud
  compression,'' \emph{IEEE Transactions on Circuits and Systems for Video
  Technology}, vol.~31, no.~2, pp. 765--781, 2020.

\bibitem{zhu2021lossy}
W.~Zhu, Y.~Xu, D.~Ding, Z.~Ma, and M.~Nilsson, ``Lossy point cloud geometry
  compression via region-wise processing,'' \emph{IEEE Transactions on Circuits
  and Systems for Video Technology}, vol.~31, no.~12, pp. 4575--4589, 2021.

\bibitem{akhtar2019low}
A.~Akhtar, B.~Kathariya, and Z.~Li, ``Low latency scalable point cloud
  communication,'' in \emph{2019 IEEE International Conference on Image
  Processing (ICIP)}.\hskip 1em plus 0.5em minus 0.4em\relax IEEE, 2019, pp.
  2369--2373.

\bibitem{zhang2021attan}
G.~Zhang, Q.~Ma, L.~Jiao, F.~Liu, and Q.~Sun, ``Attan: Attention adversarial
  networks for 3d point cloud semantic segmentation,'' in \emph{Proceedings of
  the Twenty-Ninth International Conference on International Joint Conferences
  on Artificial Intelligence}, 2021, pp. 789--796.

\bibitem{9906120}
Y.~Yu, W.~Zhang, G.~Li, and F.~Yang, ``A regularized projection-based geometry
  compression scheme for lidar point cloud,'' \emph{IEEE Transactions on
  Circuits and Systems for Video Technology}, pp. 1--1, 2022.

\bibitem{li2022frame}
L.~Li, Z.~Li, S.~Liu, and H.~Li, ``Frame-level rate control for geometry-based
  lidar point cloud compression,'' \emph{IEEE Transactions on Multimedia},
  2022.

\bibitem{houshiar20153d}
H.~Houshiar and A.~N{\"u}chter, ``3d point cloud compression using conventional
  image compression for efficient data transmission,'' in \emph{2015 XXV
  International Conference on Information, Communication and Automation
  Technologies (ICAT)}.\hskip 1em plus 0.5em minus 0.4em\relax IEEE, 2015, pp.
  1--8.

\bibitem{bentley1975multidimensional}
J.~L. Bentley, ``Multidimensional binary search trees used for associative
  searching,'' \emph{Communications of the ACM}, vol.~18, no.~9, pp. 509--517,
  1975.

\bibitem{devillers2000geometric}
O.~Devillers and P.-M. Gandoin, ``Geometric compression for interactive
  transmission,'' in \emph{Proceedings Visualization 2000. VIS 2000 (Cat. No.
  00CH37145)}.\hskip 1em plus 0.5em minus 0.4em\relax IEEE, 2000, pp. 319--326.

\bibitem{schnabel2006octree}
R.~Schnabel and R.~Klein, ``Octree-based point-cloud compression.'' \emph{PBG@
  SIGGRAPH}, vol.~3, 2006.

\bibitem{meagher1982geometric}
D.~Meagher, ``Geometric modeling using octree encoding,'' \emph{Computer
  graphics and image processing}, vol.~19, no.~2, pp. 129--147, 1982.

\bibitem{schwarz2018emerging}
S.~Schwarz, M.~Preda, V.~Baroncini, M.~Budagavi, P.~Cesar, P.~A. Chou, R.~A.
  Cohen, M.~Krivoku{\'c}a, S.~Lasserre, Z.~Li \emph{et~al.}, ``Emerging mpeg
  standards for point cloud compression,'' \emph{IEEE Journal on Emerging and
  Selected Topics in Circuits and Systems}, vol.~9, no.~1, pp. 133--148, 2018.

\bibitem{balle2017end}
J.~Ball{\'e}, V.~Laparra, and E.~P. Simoncelli, ``End-to-end optimized image
  compression,'' in \emph{5th International Conference on Learning
  Representations, ICLR 2017}, 2017.

\bibitem{balle2018variational}
J.~Ball{\'e}, D.~Minnen, S.~Singh, S.~J. Hwang, and N.~Johnston, ``Variational
  image compression with a scale hyperprior,'' in \emph{International
  Conference on Learning Representations}, 2018.

\bibitem{lu2019dvc}
G.~Lu, W.~Ouyang, D.~Xu, X.~Zhang, C.~Cai, and Z.~Gao, ``Dvc: An end-to-end
  deep video compression framework,'' in \emph{Proceedings of the IEEE/CVF
  Conference on Computer Vision and Pattern Recognition}, 2019, pp.
  11\,006--11\,015.

\bibitem{hu2021fvc}
Z.~Hu, G.~Lu, and D.~Xu, ``Fvc: A new framework towards deep video compression
  in feature space,'' in \emph{Proceedings of the IEEE/CVF Conference on
  Computer Vision and Pattern Recognition}, 2021, pp. 1502--1511.

\bibitem{hu2022coarse}
Z.~Hu, G.~Lu, J.~Guo, S.~Liu, W.~Jiang, and D.~Xu, ``Coarse-to-fine deep video
  coding with hyperprior-guided mode prediction,'' in \emph{Proceedings of the
  IEEE/CVF Conference on Computer Vision and Pattern Recognition}, 2022, pp.
  5921--5930.

\bibitem{li2021deep}
J.~Li, B.~Li, and Y.~Lu, ``Deep contextual video compression,'' \emph{Advances
  in Neural Information Processing Systems}, vol.~34, pp. 18\,114--18\,125,
  2021.

\bibitem{huang2020octsqueeze}
L.~Huang, S.~Wang, K.~Wong, J.~Liu, and R.~Urtasun, ``Octsqueeze:
  Octree-structured entropy model for lidar compression,'' in \emph{Proceedings
  of the IEEE/CVF conference on computer vision and pattern recognition}, 2020,
  pp. 1313--1323.

\bibitem{biswas2020muscle}
S.~Biswas, J.~Liu, K.~Wong, S.~Wang, and R.~Urtasun, ``Muscle: Multi sweep
  compression of lidar using deep entropy models,'' \emph{Advances in Neural
  Information Processing Systems}, vol.~33, pp. 22\,170--22\,181, 2020.

\bibitem{que2021voxelcontext}
Z.~Que, G.~Lu, and D.~Xu, ``Voxelcontext-net: An octree based framework for
  point cloud compression,'' in \emph{Proceedings of the IEEE/CVF Conference on
  Computer Vision and Pattern Recognition}, 2021, pp. 6042--6051.

\bibitem{fu2022octattention}
C.~Fu, G.~Li, R.~Song, W.~Gao, and S.~Liu, ``Octattention: Octree-based
  large-scale contexts model for point cloud compression,'' in
  \emph{Proceedings of the AAAI Conference on Artificial Intelligence}, 2022,
  pp. 625--633.

\bibitem{shannon2001mathematical}
C.~E. Shannon, ``A mathematical theory of communication,'' \emph{ACM SIGMOBILE
  mobile computing and communications review}, vol.~5, no.~1, pp. 3--55, 2001.

\bibitem{van2016pixel}
A.~Van Den~Oord, N.~Kalchbrenner, and K.~Kavukcuoglu, ``Pixel recurrent neural
  networks,'' in \emph{International conference on machine learning}.\hskip 1em
  plus 0.5em minus 0.4em\relax PMLR, 2016, pp. 1747--1756.

\bibitem{salimans2017pixelcnn++}
T.~Salimans, A.~Karpathy, X.~Chen, and D.~P. Kingma, ``Pixelcnn++: Improving
  the pixelcnn with discretized logistic mixture likelihood and other
  modifications,'' \emph{arXiv preprint arXiv:1701.05517}, 2017.

\bibitem{choy20194d}
C.~Choy, J.~Gwak, and S.~Savarese, ``4d spatio-temporal convnets: Minkowski
  convolutional neural networks,'' in \emph{Proceedings of the IEEE/CVF
  Conference on Computer Vision and Pattern Recognition}, 2019, pp. 3075--3084.

\bibitem{kathariya2018scalable}
B.~Kathariya, L.~Li, Z.~Li, J.~Alvarez, and J.~Chen, ``Scalable point cloud
  geometry coding with binary tree embedded quadtree,'' in \emph{2018 IEEE
  International Conference on Multimedia and Expo (ICME)}.\hskip 1em plus 0.5em
  minus 0.4em\relax IEEE, 2018, pp. 1--6.

\bibitem{huang20193d}
T.~Huang and Y.~Liu, ``3d point cloud geometry compression on deep learning,''
  in \emph{Proceedings of the 27th ACM International Conference on Multimedia},
  2019, pp. 890--898.

\bibitem{qi2017pointnet++}
C.~R. Qi, L.~Yi, H.~Su, and L.~J. Guibas, ``Pointnet++: Deep hierarchical
  feature learning on point sets in a metric space,'' \emph{Advances in neural
  information processing systems}, vol.~30, 2017.

\bibitem{gao2021point}
L.~Gao, T.~Fan, J.~Wang, Y.~Xu, J.~Sun, and Z.~Ma, ``Point cloud geometry
  compression via neural graph sampling,'' in \emph{2021 IEEE International
  Conference on Image Processing (ICIP)}.\hskip 1em plus 0.5em minus
  0.4em\relax IEEE, 2021, pp. 3373--3377.

\bibitem{chang2015shapenet}
A.~X. Chang, T.~Funkhouser, L.~Guibas, P.~Hanrahan, Q.~Huang, Z.~Li,
  S.~Savarese, M.~Savva, S.~Song, H.~Su \emph{et~al.}, ``Shapenet: An
  information-rich 3d model repository,'' \emph{arXiv preprint
  arXiv:1512.03012}, 2015.

\bibitem{wang2021lossy}
J.~Wang, H.~Zhu, H.~Liu, and Z.~Ma, ``Lossy point cloud geometry compression
  via end-to-end learning,'' \emph{IEEE Transactions on Circuits and Systems
  for Video Technology}, vol.~31, no.~12, pp. 4909--4923, 2021.

\bibitem{wang2021multiscale}
J.~Wang, D.~Ding, Z.~Li, and Z.~Ma, ``Multiscale point cloud geometry
  compression,'' in \emph{2021 Data Compression Conference (DCC)}.\hskip 1em
  plus 0.5em minus 0.4em\relax IEEE, 2021, pp. 73--82.

\bibitem{d20178i}
E.~d’Eon, B.~Harrison, T.~Myers, and A.~C. Philip, ``8i voxelized full
  bodies-a voxelized point cloud dataset,'' \emph{ISO/IEC JTC1/SC29 Joint
  WG11/WG1 (MPEG/JPEG) input document WG11M40059/WG1M74006}, 2017.

\bibitem{d2016mvub}
L.~Charles, C.~Qin, O.~Sergio, and P.~A.~Chou, ``Microsoft voxelized upper
  bodies - a voxelized point cloud dataset,'' \emph{ISO/IEC MPEG m38673}, 2016.

\bibitem{bishop1998latent}
C.~M. Bishop, ``Latent variable models,'' in \emph{Learning in graphical
  models}.\hskip 1em plus 0.5em minus 0.4em\relax Springer, 1998, pp. 371--403.

\bibitem{szegedy2017inception}
C.~Szegedy, S.~Ioffe, V.~Vanhoucke, and A.~A. Alemi, ``Inception-v4,
  inception-resnet and the impact of residual connections on learning,'' in
  \emph{Thirty-first AAAI conference on artificial intelligence}, 2017.

\bibitem{hu2020coarse}
Y.~Hu, W.~Yang, and J.~Liu, ``Coarse-to-fine hyper-prior modeling for learned
  image compression,'' in \emph{Proceedings of the AAAI Conference on
  Artificial Intelligence}, vol.~34, no.~07, 2020, pp. 11\,013--11\,020.

\bibitem{behley2019semantickitti}
J.~Behley, M.~Garbade, A.~Milioto, J.~Quenzel, S.~Behnke, C.~Stachniss, and
  J.~Gall, ``Semantickitti: A dataset for semantic scene understanding of lidar
  sequences,'' in \emph{Proceedings of the IEEE/CVF International Conference on
  Computer Vision}, 2019, pp. 9297--9307.

\bibitem{schwarz2018common}
S.~Schwarz, G.~Martin-Cocher, D.~Flynn, and M.~Budagavi, ``Common test
  conditions for point cloud compression,'' \emph{Document ISO/IEC
  JTC1/SC29/WG11 w17766, Ljubljana, Slovenia}, 2018.

\bibitem{muh2022vox}
M.~Asad~Lodhi, J.~Pang, and D.~Tian, ``Point cloud geometry compression using
  learned octree entropy coding,'' \emph{ISO/IEC JTC1/SC29/WG7 m59528}, 2022.

\bibitem{bingham2001random}
E.~Bingham and H.~Mannila, ``Random projection in dimensionality reduction:
  applications to image and text data,'' in \emph{Proceedings of the seventh
  ACM SIGKDD international conference on Knowledge discovery and data mining},
  2001, pp. 245--250.

\end{thebibliography}

%








\end{document}